\documentclass{article}

\usepackage{arxiv}

\usepackage[utf8]{inputenc} % allow utf-8 input
\usepackage[T1]{fontenc}    % use 8-bit T1 fonts
\usepackage{hyperref}       % hyperlinks
\usepackage{url}            % simple URL typesetting
\usepackage{booktabs}       % professional-quality tables
\usepackage{amsfonts}       % blackboard math symbols
\usepackage{nicefrac}       % compact symbols for 1/2, etc.
\usepackage{microtype}      % microtypography
\usepackage{graphicx}
\usepackage{natbib}
\usepackage{doi}
\usepackage{amsthm}
\usepackage{amsmath}
\usepackage{svg}

\title{COGNITIVE MODELING AND LEARNING \\ WITH SPARSE BINARY HYPERVECTORS}

\author{
\hspace{1mm}Zhonghao Yang\href{https://orcid.org/0009-0003-6353-9150}{\includegraphics[scale=0.06]{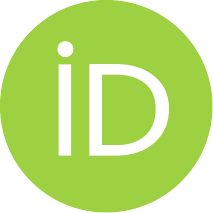}}\\
\\
Independent Researcher\\
\texttt{yangzh@gmail.com} \\
}

\renewcommand{\headeright}{Technical Report}
\renewcommand{\undertitle}{Technical Report}
\renewcommand{\shorttitle}{Cognitive Modeling and Learning with Sparse Binary Hypervectors}

\hypersetup{
pdftitle={cognitive modeling and learning with sparse binary hypervectors},
pdfauthor={Zhonghao Yang},
pdfkeywords={Artificial Intelligence, Vector Symbolic Architecture, VSA, Hyper-dimensional Computing},
}

\begin{document}
\maketitle

\begin{abstract}
 Following the general theoretical framework of VSA (Vector Symbolic Architecture), a cognitive model with the use of sparse binary hypervectors is proposed, In addition, learning algorithms are introduced to learn / bootstrap the model from incoming data stream, with much improved transparency and efficiency. Mimicking human cognitive process, the model's training can be performed online while inference is in session. Word-level embedding is re-visited with such hypervectors, and further applications in the field of NLP (Natural Language Processing) are explored. 
\end{abstract}


\section{Introduction}

Deep neural networks are AI models that are used extensively in recent years with growing popularity. However, they suffer several notable deficiencies from its original design since early history, which the academia and the industry have spent millions of dollars trying to address with varying degree of success.

\textbf{Model transparency / explainability}: deep neural networks are generally considered as black boxes where the inside is hard to discern, let alone make necessary adjustments when mismatch occurs. There do exist, primarily in academia, efforts to render the model more transparent and thus the decision-making more equitable, however, there is no well-accepted solution so far for general deep neural networks.

Depending on the real-world applications in question, especially when the consequence of failure is trivial, the lack of transparency can be a non-issue. For example, nobody will throw away his phone if the image search on the phone missed one photo of a bunny. However, for mission-critical applications such as autonomous driving, industrial assembly line, power grid / energy infrastructure controller, we need to clearly understand how the model operates, and quantify possible failure modes, if any.

The argument also applies to applications in various business domains. For example, an AI model taking actions against a social media post (or a potential malicious app) will need a clear justification on why the decision was made, for upcoming review or potential legal challenge.

Note the expected justifications can be either part of the model output, in the similar fashion as the \textit{train of thoughts} in current ChatGPT service \footnote{\url{https://chat.openai.com}} whereas the underlying model is opaque, or can be produced directly from the internal state of the model. While the former approach can certainly help, the publication will focus on the latter.

Modern neural networks are often plagued by hallucination, with which the model produces unexpected output (inappropriate, plainly absurd, or even worse, harmful), a manifestation that we have insufficient understanding and control over the underlying model. Again, the industry is struggling with an effective solution, as it seems to be an inherent drawback of deep neural networks.

\textbf{Cost}: this generally refers to the computation cost, loosely related to the raw power consumption, the CPUs / GPUs needed (especially during training), and the labor cost.

A typical training for ChatGPT models costs millions of dollars, several months of continuous GPU hours, and many highly-trained engineers with sophisticated process, making modern deep neural networks a monopoly / privilege of those companies with deep pockets. In contrast, this publication aims at solutions that can reduce the overall cost significantly, even orders of magnitude, making viable efforts into AI democratization.

\textbf{Efficiency}: this is closely related to cost. If the training and deployment of AI models are expensive, everything has to be optimized around cost considerations. Reduced cost can lead to greater efficiency in terms of organizational processes.

Furthermore, on the representation level, the state-of-the-art deep neural networks contains millions or even billions of floating-point weights. Moving these huge amount of weights between external storage, memory, CPU and GPU posts great burden for the underlying infrastructure such as high-bandwidth and low-latency inter-connections. High efficient solutions, in principle, can dramatically reduce the requirement for supporting systems and infrastructure.

On the algorithm level, modern deep neural networks typically performs training via back propagation, which performs gradual tweaking over millions of weights by repeatedly iterating over training sets. With Internet-scale training set, updating huge set of weights can be extremely bulky and laborious, even with the latest development of GPUs and large on-board memory and cache. Any improvement at learning algorithm can have a huge impact on the overall efficiency of AI models.

The mobile devices (and the Internet-of-Things devices) have limited capacity in terms of storage and computing, and severely power-constrained, which makes the deployment of deep neural networks quite challenging. Again, high-efficient AI models (and learning algorithms) that require significantly less storage and computation will be highly desirable for deployment to these devices.

Initially inspired by human brain and its unique cognitive capabilities, VSA (Vector Symbolic Architecture, \cite{Gayler2004vector}) or HDC (Hyper-Dimensional Computing) \cite{Kenerva2009} is another school of AI models. While it shares a few traits with connectionist models such as neural networks, it's better (and more properly) characterized as a symbolic approach in the heated debate between the dichotomy of symbolic and connectionist AI. The functioning units (of hypervectors from VSA models) collectively represent unique meaning (or symbols), and it's algebraic operations between these hypervectors that models the interactions of real-world entities. 

Unlike the well-known mantra of "curse of dimensionality" in machine learning community where high-dimensional space is typically frown upon, VSA takes advantage of the high dimensionality in hypervectors, dubbed as "blessing of dimensionality". The high-dimensional space offers quite unique mathematical and topological properties that enabled and empowered the breakthrough outlined here. \cite{Kenerva2009} is highly recommended for interested readers.

VSA has witnessed growing popularity in academia and some industrial settings in recent years. However, its usage is mostly exploratory, due to a few critical questions left unaddressed:
\begin{itemize}
    \item On the theoretical level, what construction of the model offers the most benefit. What unique and appealing capability does this particular embodiment provide?
    \item What do the overall learning algorithms (or learners) look like?
    \item Compared with deep neural networks, what is use case, that demonstrates the most clear benefit and advantage?
\end{itemize}

\section{Cognitive Principles}

\subsection{Sparse binary hypervectors}

Unlike fore-mentioned VSA, which is mostly a theoretical framework accommodating wide range of configuration choices, the publication fixates on the use of sparse binary hypervectors.

Define \textbf{sparsity} $s=M/N$ as the fraction of ON bits count $M$ over the dimension $N$ (of an binary vector). 

Being sparse implies the count of ON bits is relatively small, thus looking sparse among $N$ dimensions. Sparse binary hypervectors are merely binary vectors with large dimension $N$ and low sparsity $s$, typically $N \gg 1000$ and $s \ll 0.01$.  

In the following context, $N=2^{16}=65536$, $s=1/256$ (and thus $M=256$) are used. All such hypervectors forms the space of $\mathbb{C}$. Obviously the general principles about hypervectors apply to other configurations of $N$ and $s$.

\begin{figure}
  \centering
  \includegraphics[scale=1]{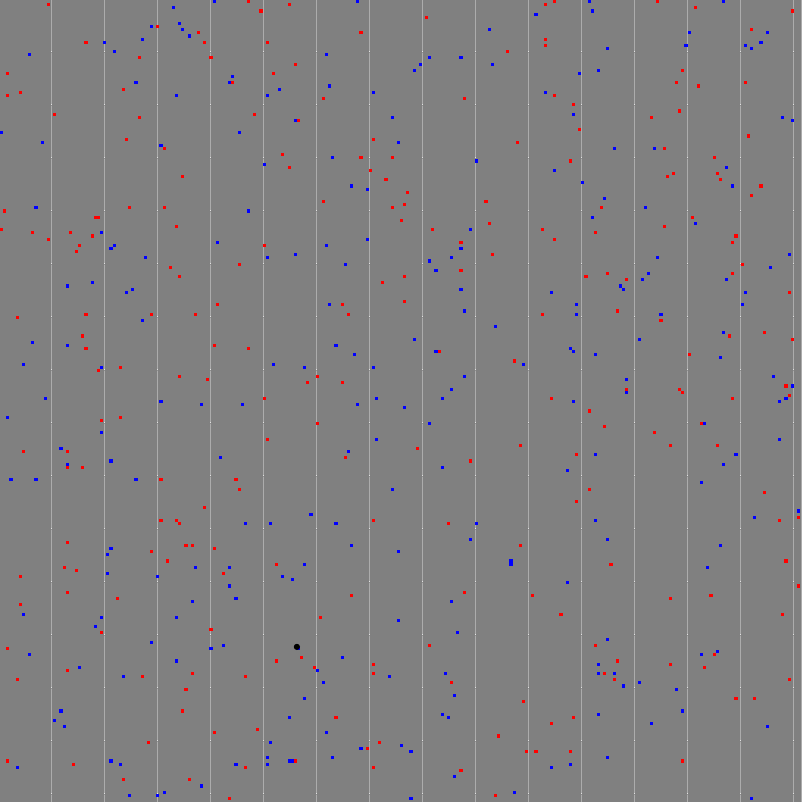} 
  \caption{ON bits for two random hypervectors from $\mathbb{C}$}
  \label{fig:hypervectors}
\end{figure}

FIG.\ref{fig:hypervectors} offers an intuitive glimpse of two random hypervectors in $\mathbb{C}$, one by red dots and another by blue. Each set (of a single hypervector) of $M=256$ ON bits is spread among the $16 \times 16$ cells, where each cell contains $16 \times 16$ bit positions, totalling $N=65536$. If you examine closely, there is a single fat dot in black, marking the single (and very unlikely) overlap between them. For random hypervectors, the similarity is minimal and any overlap is extremely unlikely.

Within this space of $\mathbb{C}$, \textbf{inner product} $\langle A, B \rangle$ is introduced. For any $A$, $\| A \|^2 = \langle A, A \rangle = M$: all vectors have the same length of $\sqrt{M}$.

The \textit{alignment} between their vector forms in $\mathbb{C}$ tells the correlation, that is, how similar these two hypervectors are. For any $A$ and $B$, their cosine angle (between corresponding vectors $\vec{OA}$ and $\vec{OB}$) is simply $\frac{\langle A, B \rangle}{M}$.

\subsection{Overlap and semantic similarity}

Define \textbf{overlap} for two hypervectors $A$ and $B$ (from $\mathbb{C}$) as the count of ON bits where $A_i = B_i = 1$, for $i \in [0, N)$. This is equivalent to the inner product $\langle A, B \rangle$, also bitwise AND operations for binary hypervectors.
\begin{equation}
O(A, B)=\sum^{N} {A_i}{B_i}=\langle A, B \rangle \label{eq:overlap_def}
\end{equation}

Define \textbf{Hamming distance} $H(A, B)$ for two hypervectors $A$ and $B$ (from $\mathbb{C}$), which is equivalent to bitwise XOR operations for binary hypervectors.

With identical hypervectors having overlap of $Ns=M$ as one extreme, and a pair of hypervectors (of complete randomness) has the overlap of $Ns^2=Ms$, as another extreme, the ratio of $1/s$ can be thought as signal-to-noise ratio (SNR) for those readers with electrical engineering background: the smaller $s$ becomes, the large disparity between these two extremes. The desirable high SNR seems to be another strong reason to support the use of sparse hypervectors.

For arbitrary hypervector $A$ and $B$ in the space of $\mathbb{C}$, we have
\begin{equation}
2 \times O(A, B) + H(A, B) = 2M \label{eq:overlap_hamming}
\end{equation}

\begin{proof}
Suppose $A^*$ is the set of ON positions (indices) in $A$, and $B^*$ is the set of ON positions (indices) in $B$.

We have
\begin{align*}
|A^* \cup B^*| &= |A^*| + |B^*| - |A^* \cap B^*| \\
|A^* \cup B^*| &= |A^* \cap B^*| + |A^* \setminus B^*| + |B^* \setminus A^*|
\end{align*}
where $|A^*|$ denotes the cardinality of the set $A^*$.

Note that
$|A^* \setminus B^*| + |B^* \setminus A^*| = H(A, B)$, $|A^* \cap B^*| = O(A,B)$,
and $|A^*|=|B^*|=M$, we thus have our proof.
\end{proof}

To sum up, Hamming distance (as a distance measure) is inversely related to overlap (as a similarity measure): they are really the two sides of the same coin that reveals the same inherent traits for hypervectors in the space of $\mathbb{C}$.

\subsection{Segmented hypervectors}

Segmented hypervectors form a subset within $\mathbb{C}$, where the total $N$ dimensions are divided into $M=Ns$ continuous segments: within each segment (of dimension $1/s$), there is one and only one ON bit. 

For the case of $N=65536$ and $s=1/256$, every $256$ dimensions form a segment, which contains one and only one ON bit. Altogether there are $M=256$ segments, and thus $256$ ON bits. The space of all segmented hypervectors is denoted as $\mathbb{C'}$: $\mathbb{C'} \subset \mathbb{C}$.

This space of $\mathbb{C'}$ has $(1/s)^{Ns}$ points, which is significantly smaller than $\mathbb{C}$. However, for all practical applications, it's still huge and seemingly unlimited, which can be verified by interested readers.

Hypervectors with $N=16$ and $s=1/4$ are demonstrated here:
\begin{equation*}
    \begin{split}
            A &= \; 0011 \; 0000 \; 1001 \; 0000 \in \mathbb{C} \; (however A \notin \mathbb{C'}) \\
            C &= \; 0010 \; 1000 \; 0001 \; 0001 \; = (2, 0, 3, 3) \; \in \mathbb{C'} \subset \mathbb{C} \\
            D &= \; 0010 \; 0100 \; 0001 \; 0100 \; = (2, 1, 3, 1) \; \in \mathbb{C'} \subset \mathbb{C} \\
      O(C, D) &= \lvert 0010 \; 0000 \; 0001 \; 0000 \;\rvert = 2 \\ 
      H(C, D) &= \lvert 0000 \; 1100 \; 0000 \; 1100 \;\rvert = 4
\end{split}
\end{equation*}

Also notice hypervectors from $\mathbb{C'}$ can be represented much more compactly by their offsets within the segments. Alternatively a hypervector from $\mathbb{C'}$ can be imaged as a slot machine with $M=Ns$ spinning wheels, where each spinning wheel has $1/s$ slots/teeth.

The representation of hypervectors of $\mathbb{C'}$ suits nicely with modern computer architecture. First of all, the hypervector itself is sparse and only non-zero bits need to be stored. Secondly, the segmented structure implies we only need to store local offsets with smaller dynamic range of $1/s$. Finally, the use of binary value instead of real-value neural network weights. Overall, one of our hypervectors ($N=65536$) will take $256$ bytes (which reflects exactly the entropy of $2048$ bits), the the same storage budget as a 64-dimensional floating-point vector: a typical vector in neural networks is much wider than 64 dimension (in the order of thousands). In addition, we don’t need floating-point operations which can be orders of magnitude slower and power hungrier than simple bits flipping.

From now on, we limit our future discussion within the space of $\mathbb{C'}$, unless stated otherwise.

A random segmented hypervector can be generated by randomly picking offsets for each segments. As part of the "blessing of high dimensionality", any pair of random hypervectors in $\mathbb{C'}$ is maximally dissimilar (almost no overlap) by construction, also known as near-orthogonal in mathematical jargon.

For cognitive entities range from concrete objects (such as people, physical objects) to abstract concepts (such as ideas, alphabets, words, novels), and everything in between, we boldly hypothesize that all of them can be modeled with sparse hypervectors from $\mathbb{C'}$. It's the algebraic interactions that we will outline next that mirrors the interactions between cognitive entities in a world model.  

\subsection{Bundle}

Define \textbf{bundle} for $K$ random codes $C_k$ (from $\mathbb{C'}$) with normalized weights $w_k$ ($\sum_K w_k = 1$) as
\begin{equation}
B= (w_0 \cdot C_0) \oplus (w_1 \cdot C_1) \oplus ... \oplus (w_{K-1} \cdot C_{K-1})\label{eq:bundle}
\end{equation}
with the ON offset at each segment $B_i$ ($0 \le i < M$) is probabilistically determined by reusing offset from $C_{k,i}$ with the probability of $w_k$. 

The resultant $B$ remains in $\mathbb{C'}$, and is similar to any of its operands $C_k$, due to construction:
\begin{equation}
O(B, C_k) = \langle B, C_k \rangle \approx w_k Ns\label{eq:bundle_overlap}
\end{equation}

Think this as an unique lossy compression scheme, $B$ maximally retains the original segment-wise offsets from codes $C_k$, proportional to its weight $w_k$.

For a special case where $w_k=1/K$ for all $k$, \textbf{bundle} notation can be simplified as 
\begin{equation}
B = C_0 \oplus C_1 ... \oplus C_{K-1}
\end{equation}
and consequently
\begin{equation}
    O(C_k, B)=\langle C_k, B \rangle \approx \frac{1}{K} Ns
\end{equation}
but keep in mind this is only a special case, and the weights of $1/K$ are implicit.

The \textbf{bundle} operation is commutative:
\begin{align*}
    A \oplus B &= B \oplus A
\end{align*}

\subsubsection{Conformants}

Let's take a closer look at the bundle operation Eq.\eqref{eq:bundle}: it's non-deterministic in the sense that random generators with different seeds will produce different results by picking segments differently.

All possible results forms a subspace $\tilde B$, where each and every member equally conforms to Eq.\eqref{eq:bundle_overlap}: the collection of results are thus called \textbf{conformants}.

As we experience (and encode) the same set of $\{C_k\}$ , which can result in distinct conformants $B^{(0)}$, $B^{(1)}$ in $\tilde B$. 

Obviously 
\begin{align*}
    O(B^{(0)}, C_k) = O(B^{(1)}, C_k) \approx \frac{1}{K}Ns
\end{align*}

However, in addition we will have
\begin{equation}
    O(B^{(0)}, B^{(1)}) = \frac{1}{K}Ns
\end{equation}

\begin{proof}
From any input $C_k$ (fix $k$ for now), $B^{(0)}$ will take $Ns/K$ ON bits, and $B^{(1)}$ will independently take another set of $Ns/K$ ON bits. On average, the overlap between $B^{(0)}$ and $B^{(1)}$ due to this fixed member $C_k$ is $Ns/K^2$.

Repeat this for every member $C_k$, we will approximately have
\begin{align*}
    O(B^{(0)}, B^{(1)}) \approx \sum_K Ns/K^2 = Ns/K
\end{align*}
\end{proof}

Overall, we think the very existence of conformants for bundle operation is an important feature instead of a design flaw, as it exposes another degree of freedom, thanks to the blessing of dimensionality.

\subsubsection{Online bundling}

Beyond representing individual hypervectors, the \textbf{online learner} $L^{(k)}$, for a data stream of incoming $C_k$ is defined as:
\begin{equation}
\begin{split}
    L^{(0)} &= C_0 \\
    L^{(1)} &= \sum_{k=0, \oplus}^{1} C_k = \frac{1}{2} C_0 \oplus \frac{1}{2} C_1 = \frac{1}{2} L^{(0)} \oplus \frac{1}{2} C_1 \\
    L^{(2)} &= \sum_{k=0, \oplus}^{2} C_k = \frac{1}{3} C_0 \oplus \frac{1}{3} C_1 \oplus \frac{1}{3} C_2 = \frac{2}{3} L^{(1)} \oplus \frac{1}{3} C_2 \\
    ... \\
    L^{(k)} &= \sum_{k=0, \oplus}^{k} C_k = \frac{1}{k+1} C_0 \oplus \frac{1}{k+1} C_1 \oplus ... \oplus \frac{1}{k+1} C_{k} = \frac{k}{k+1} L^{(k-1)} \oplus \frac{1}{k+1} C_{k}   \\
\end{split}
\end{equation}

Image the learner at an arbitrary time $k$: $L^{(k)} = \frac{k}{k+1} L^{(k-1)} \oplus \frac{1}{k+1} C_k$, we simply nudge the existing learner $L^{(k-1)}$ towards the incoming $C_k$, with the learning rate of $\frac{1}{k+1}$. This small-step of tweaking will ensure the updated learner $L^{(k)}$ will keep almost equal similarity and distance (per Eq.\eqref{eq:overlap_hamming}) to all its experiences $\{ C_k \}$ so far: the learner $L^{(k)}$ is effectively a running averaging operator of all its experiences.

In practice, the decreasing learning rate $\frac{1}{k+1}$ can be interpreted as the fraction of segments that needs updating. As time goes by, the work needs to be done by the learner (from new data point) becomes less and less: it's quite hard to dramatically change a well-experienced learner, in agreement with our daily cognitive experience.

\subsection{Bind}

Define \textbf{bind} operation of $K$ codes $C_k$ (from $\mathbb{C'}$) as 
\begin{equation}
B = C_0 \otimes C_1 \otimes ... \otimes C_{K-1}\label{eq:bind}
\end{equation}
with the offset at segment $i$ as
\begin{equation}
    B_i = (\sum_k C_{k,i}) \mod{M}
\end{equation}
where $C_{k,i}$ is the offset from $i$th segment of code $C_k$.

The resulted $B$ is maximally dissimilar to all its operands $C_k$:
\begin{align*}
    O(B, C_k) = \langle B, C_k \rangle \approx Ns^2 = 1
\end{align*}

The \textbf{bind} operation is obviously commutative and associative: 
\begin{equation}
\begin{split}
A \otimes B &= B \otimes A \\
(A \otimes B) \otimes C &= A \otimes (B \otimes C)    
\end{split}
\end{equation}

Notably, \textbf{bind} preserves \textbf{overlap} and \textbf{Hamming} distance:
\begin{equation}
\begin{split}
    O(A, B) &= O(A \otimes P, B \otimes P) \\
    H(A, B) &= H(A \otimes P, B \otimes P)
\end{split}
\end{equation}

Lastly, \textbf{bind} operation distributes over \textbf{bundle}:
\begin{equation}
    P \otimes (A \oplus B)  = P \otimes A \oplus P \otimes B,
\end{equation}
which is similar to arithmetic counterparts of $*$ and $+$. 

\subsubsection{Unit and inverse vector}

Define a \textbf{unit vector} $I$ (from $\mathbb{C'}$), such that any code $C$
\begin{equation}
    C = C \otimes I
\end{equation}

The unit vector $I$ is simply a hypervector where all the segments has ON bits at offset $0$. For example when $s=1/4$, the unit vector is $1000 \; 1000 \; 1000 \; 1000$, or conveniently recorded as $(0,0,0,0)$.

Define \textbf{inverse vector} $C^{-1}$ for a code $C$, such that 
\begin{equation}
    C^{-1} \otimes C = I    
\end{equation}
For any code $C$, the unique inverse $C^{-1}$ exists.

Actually, with the definition of \textbf{unit vector} (and \textbf{inverse vector} above), this forms an \textbf{algebraic ring} with bind and bundle operations.

\subsection{Analogical reasoning}

Analogical reasoning is perhaps the most fascinating features for VSA and hypervectors.

Define \textbf{release} operation as 
\begin{equation}
  A \oslash B = A \otimes B^{-1}    
\end{equation}

It releases / unbinds previously bound code: ($A \otimes B) \oslash B = A$.\\

We have:
\begin{align*}
  Q = (P_1 \otimes A \oplus P_2 \otimes B) \oslash P_1 \approx A + noise
\end{align*}

In practice, a near-neighbor search with $Q$ will retrieve $A$, as $O(Q, A) \approx \frac{1}{2}M$, which should be the very dominant among all recorded patterns.

Similarly
\begin{align*}
  &(P_1 \otimes A \oplus P_2 \otimes B) \oslash P_2 \approx B + noise\\
  &(P_1 \otimes A \oplus P_2 \otimes B) \oslash A \approx P_1 + noise\\
  &(P_1 \otimes A \oplus P_2 \otimes B) \oslash B \approx P_2 + noise
\end{align*}
if $A$, $B$, $P_1$, $P_2$ are all random hypervectors.\\

We will use Pentti Kanerva's \textit{the dollar of Mexico} (\cite{Kenerva2009}) as a concrete example here. Suppose we want to encode the following tabular knowledge
\begin{itemize}
  \item Mexico have the country code of MEX, capital of Mexico City and the currency of Peso;
  \item The United States have the country code of USA, capital of Washington DC, and the currency of dollar;
\end{itemize}

with these
\begin{align*}
    C_{mexico} &= P_{code} \otimes C_{mex} \oplus P_{capital} \otimes C_{mexicoCity} \oplus P_{currency} \otimes C_{peso} \\
    C_{us} &= P_{code} \otimes C_{usa} \oplus P_{capital} \otimes C_{dc} \oplus P_{currency} \otimes C_{dollar}
\end{align*}
where all elements are random hypervectors.

Retrieve individual filler given known role/attribute is straightforward:

\textbf{capital of Mexico}: 
\begin{align*}
    C_{mexico} \oslash P_{capital} \approx C_{mexicoCity} + noise
\end{align*}
\textbf{currency of United States}: 
\begin{align*}
    C_{us} \oslash P_{currency} \approx C_{dollar} + noise
\end{align*}

It's also possible to retrieve the role/attribute given known filler:
\textbf{what's the role of peso}: 
\begin{align*}
    C_{mexico} \oslash C_{peso} \approx P_{currency} + noise
\end{align*}
\textbf{what does "USA" stand for?}
\begin{align*}
    C_{us} \oslash C_{usa} \approx P_{code} + noise
\end{align*}

Even more interestingly, without knowing the attribute for a given filler, analogical reasoning can be performed.
\textbf{what's the dollar of Mexico}: 
\begin{align*}
    C_{dollar} \otimes C_{mexico} \oslash C_{us} \approx C_{peso} + noise
\end{align*}

Similarly, \textbf{what's the counterpart in Mexico as DC in US}: 
\begin{align*}
    C_{dc} \otimes C_{mexico} \oslash C_{us} &\approx C_{mexicoCity} + noise
\end{align*}
\textbf{What's to Mexico is similar to the label USA as to United States}: 
\begin{align*}
    C_{usa} \otimes C_{mexico} \oslash C_{us} &\approx C_{mex} + noise
\end{align*}

\textbf{Knowledge transfer} without decoupling first: if we construct
\begin{align*}
    C'_{us} = C_{mexico} \otimes (C_{usa} \oslash C_{mex} \oplus C_{dc} \oslash C_{mexicoCity} \oplus C_{dollar} \oslash C_{peso})
\end{align*}
then \begin{align*}
    C'_{us} \oslash P_{code} &\approx C_{usa} \\
    C'_{us} \oslash P_{capital} &\approx C_{dc} \\
    C'_{us} \oslash P_{currency} &\approx C_{dollar}
\end{align*}

The proofs will be left for interested readers.

\subsection{Discussions}

\cite{Kenerva2009} kick-started the concept of hyper-dimensional computing with the use of algebraic operations upon hypervectors.
This line of research dates back to his seminal work (\cite{Kenerva1988}) in the 1980s.

\cite{Kleyko2020} and \cite{Schlegel2020} provide excellent surveys the VSA for the past 30 years. In general and not coincidentally, sparsity plays a critical role in terms of memory capacity. In addition, \cite{kleyko2018} discusses the choice of sparsity for hypervectors with experimental evidence: it turns out sparse hypervectors can achieve desirable performance, and with neural plausibility. 

A separate earlier publication \cite{mika2015} shares idea with this publication. However, we present original and critical steps forward, such as the formalization of segmented hypervectors, the novel bundle and bind operations, online bundling learner, with additional real-world application discussions. 

In addition to the theoretical foundation for a new cognitive model as presented in this section, three design principles for intelligent systems seem to emerge:
\begin{enumerate}
    \item Cognitive entities can be modeled by sparse binary hypervectors, for example, from $\mathbb{C'}$;
    \item Overlap (of sparse binary hypervectors) is a good measurement of semantic similarity. Furthermore, overlap and Hamming distance reflect the same inherent traits among cognitive entities;
    \item Compositional structures need to be encoded in the same high-dimensional space recursively ($\mathbb{C'}$, for example), which will be discussed in the next section;
\end{enumerate}

The proposed cognitive model, together with these design principles can be realized with software systems, as well as a number of hardware architectures, possibly with different materials and substrates. 

\section{Compositional structures}

\subsection{Nearly orthogonal sets}

Define an \textit{nearly orthogonal set} (NOS) as a set of codes $\{ A_i \}$ (all from $\mathbb{C'}$), for any member $A_i$ and $A_j$ ($i \ne j$):
\begin{equation}
\begin{split}
\langle A_i, A_i \rangle &= O(A_i, A_i) = Ns \\
\langle A_i, A_j \rangle &= O(A_i, A_j) = Ns^2 \approx 0 \\
\end{split}    
\end{equation}

The word of "nearly" refers to the fact that the cross inner product is only \textit{approximately} zero, with $Ns^2$ being the inherent noise.

Occasionally (but equivalently) we use \textit{relative overlap} for an nearly orthogonal set.
\begin{equation}
\begin{split}
RO(A_i,A_i) &= \langle A_i, A_i \rangle / M = 1 \\ 
RO(A_i,A_j) &= \langle A_i, A_j \rangle / M = s \approx 0   
\end{split}    
\end{equation}

\subsection{Sets}

A \textit{set} (as denoted by $\{C_k\}$) is formed implicitly from a \textit{nearly orthogonal set}, with no meaningful ordering.

The composite code $S$ for the whole set $\{ C_k \}$ can be constructed as:
\begin{equation}
S= C_0 \oplus C_1 \oplus ... \oplus C_{K-1} = \sum_{K, \oplus} C_k
\label{eq:set}
\end{equation}

The near-neighbor search with a probe of $S$ will eventually yield all member $C_k$, as 
\begin{equation}
O(S, C_k) \approx Ns/K     
\label{eq:set_overlap}
\end{equation}
for all possible $k$. The set cardinality $K$ can be recovered by counting all recovered codes.

$S$ is the centroid for the whole cluster of $\{ C_k \}$, as geometrically $S$ has approximately equal distance to all the cluster members $C_k$, thanks to Eq.\eqref{eq:overlap_hamming}. $S$ can be also considered as a summary (or a compressed version) for the whole set of codes $\{C_k\}$.

\subsection{Sequences}

A \textit{sequence} (as denoted by $[C_k]$) is formed from a \textit{nearly orthogonal set}, with enforced ordering. 

The sequences can be encoded similarly, with the additional $P_k$ as the positional markers:
\begin{equation}
S= C_0 \otimes P_0 \oplus C_1 \otimes P_1 \oplus ... \oplus C_{K-1} \otimes P_{K-1} = \sum_{K, \oplus} (C_k \otimes P_k)
\end{equation}

As we explained before, $S$ is similar to the permutated member $C_k$. A near-neighbor search with a probe of $S \oslash P_k$ (or equivalently $S \times P_k^{-1}$) can recover the member at a particular position $k$, since

\begin{equation}
O(S, C_k \otimes P_k) = O(S \oslash P_k, C_k) \approx Ns/K
\end{equation}

For further simplification, we use $P_k = P^{k}_{step}$, where $P_{step}$ is a well-known positional marker, in this case, 

\begin{equation}
    S = C_0 \otimes P^{0}_{step} \oplus C_1 \otimes P^{1}_{step} \oplus ... \oplus C_{K-1} \otimes P^{K-1}_{step} = \sum_{K, \oplus} (C_k \otimes P^{k}_{step})
\label{eq:sequence}
\end{equation}

The $k$th element can be recovered, as
\begin{equation}
O(S, C_k \otimes P_{step}^k) = O(S \oslash P_{step}^{k}, C_{k}) \approx Ns/K
\end{equation}

The near-neighbor search (with probe of $S \oslash P_k$) can retrieve the $k$th member of $C_k$: we progress with positional marker $P_k$, until no similar code can be found. 

This may remind some readers of \textit{positional encoding}, as introduced in transformer architecture (\cite{Vaswani2017}): our encoding and retrieval scheme seems to be much cleaner.

We cannot emphasis more about the importance of an efficient near-neighbor search module for this model. In this context, the module will be specifically tuned for sparse binary hypervectors.

\subsection{A Probabilistic Interpretation}
Assume we have two bundling operations, based on the same \textit{nearly orthogonal set} $\{P_k\}$:
\begin{equation}
\begin{split}
    A &=\sum_{K, \oplus} \alpha_k P_k, \; \sum_K \alpha_k = 1 \\
    B &=\sum_{K, \oplus} \beta_k P_k, \; \sum_K \beta_k = 1    
\end{split}
\end{equation}

then
\begin{equation}
\begin{split}
    \langle A, B \rangle &= \sum_K {\alpha_k \beta_k \langle C_k, C_k \rangle} + \sum_{i \ne j} {\alpha_i \beta_j \langle C_i, C_j \rangle}\\
    &= (\sum_K {\alpha_k \beta_k}) Ns + (\sum_K \alpha_k \sum_K \beta_k -\sum_K {\alpha_k \beta_k})Ns^2\\
    &= (\sum_K {\alpha_k \beta_k}) Ns + (1 -\sum_K {\alpha_k \beta_k})Ns^2\\
    &=Ns(1-s)\sum_K {\alpha_k \beta_k} + Ns^2
\end{split}
\end{equation}

The similarity between two bundled hypervectors is dominated by coefficients from orthogonal terms, especially when sparsity is low.

Define \textbf{frame inner product} as the inner product of the coefficients, under the shared frame of NOS $\{P_k\}$:
\begin{equation}
\langle A, B \rangle^* = \sum_K {\alpha_k \beta_k} = \frac{\langle A, B \rangle}{Ns(1-s)} - \frac{s}{1-s}
\label{eq:frame_inner_product}
\end{equation}

Image if the set of $P_k$ is complete orthogonal to each other: $\langle P_k, P_k \rangle = Ns$, for any $k \in [0, K)$, and $\langle P_i, P_j \rangle = 0$, for any $i \ne j$, the computation of frame inner product will be straightforward and uninteresting:

\begin{equation*}
    \langle A, B \rangle^{*} = \frac{\langle A, B \rangle}{Ns}
\label{eq:frame_inner_product_orthogonal}
\end{equation*}

The last term $\frac{s}{1-s}$ can be considered as an inherent system bias when sparsity $s$ is present.

When we use $\beta_{k}=1$ and $\beta_{i}=0$ for all $i \ne k$, $B$ becomes the probe $P_k$, and 

\begin{equation}
\alpha_k = \frac{\langle A, P_k \rangle}{Ns(1-s)} - \frac{s}{1-s} = p(P_k|A, \{P_k\})
\label{eq:empirical_count}
\end{equation}

$\alpha_k$ ($0 \le \alpha_k < 1$) is actually the empirical probability of member $P_k$ within experience recorded by $A$. 

Suppose we have an online learner $L^{(t)} = \sum_{t, \oplus} P_t$, where $P_t$ picked from a fixed \textit{NOS} of $\{ P_k \}$, duplication is not only possible, but frequent as well. Eventually it will contain the empirical probabilities associated with each and every member $P_k$. Effectively the learner itself can be considered as a probability mess function, or a probability profile. 

From the opposite direction, with the probability mess function $W_k$, it's trivial to encode:

\begin{equation}
    C = \sum_{K, \oplus} w_k P_k
\end{equation}

We can draw an nice analogy to Fourier transformation, which decomposes an arbitrary function into a set of coefficients with sinuous functions of different frequencies, phases and amplitudes. Similarly, the snapshot of an online learner corresponds to a point in the $K$-dimensional feature space, spanned by the NOS: in reality, the unknown $K$ can be in the neighborhood of thousands, but typically $K \ll N$.

\section{Applications}

\subsection{Word-level embedding}
The idea of word-level embedding has its long history in NLP (Natural Language Processing). Proposed in the late 1990s, the idea is to model the semantic meaning of each word (for example, from English) as a long vector (dimensionality typically in hundreds), which is almost always real-valued. Modern NLP systems typically use dimension around thousands.

Distributional hypothesis claims that the semantics of a word is determined by its neighbor (or contexts). With a large amount of training corpus (for example, Wikipedia or even the whole Internet), any words which have the similar contexts will have similar word embeddings, and presumably similar semantic meaning. Despite certain idiosyncrasies, this hypothesis, for the most part, inspired projects such as Word2Vec(\cite{mikolov2013efficient}, \cite{Mikolov2013}) or GloVe (\cite{Pennington2014}), which lays the foundations for the most recent deep-learning-based NLP resurgence in applications such as text summarization, machine translation and chatbots.

Acquisition of word-level embedding is traditionally done by neural networks. The weights are adjusted by back propagation in such way that they will better predict the neighboring context words. The training itself is self-supervised in the sense the training corpus provides all necessary "labels" for learning. Since the training itself is still non-trivial, with the Internet-scale data set, most companies simply use the pre-trained embeddings and focus on their own downstream applications.

We believe the algebraic operations of \textbf{bind} and \textbf{bundling} in $\mathbb{C'}$ and the online learner can deliver significant benefits with improved transparency and efficiency, which inspired us to revisit this problem. 

We use a half window size of $2$ around a center word, respectively $c_{-2}, c_{-1}, w, c_1, c_2$. The window size can be trivially expanded when needed.

For this occurrence $t$ of word $w$ in the training corpus, an observation hypervector is produced:
\begin{align}
    C^{(t)}_{w} = C_{-2} \otimes P_{step}^{-2} \oplus C_{-1} \otimes P_{step}^{-1} \oplus C_w \oplus C_{1} \otimes P_{step} \oplus C_{2} \otimes P_{step}^2
\end{align}
where $C_{-2}$, $C_{-1}$, $C_1$, $C_2$ are the codes for context words $c_{-2}$, $c_{-1}$, $c_1$ and $c_2$, and $C_w$ the code for the center word $w$. $P_{step}$ is a well-known step marker.

A document with $W$ words will produce $W$ such observation hypervectors, aiming at different center word $w$.

The learner for a fixed center word $w$ follows:
\begin{align}
    L^{(T)}_w = \sum_{T, \oplus} C^{(t)}_{w}
\end{align}

We maintain one learner $L_w$ for each word $w$ throughout the training, fed by the observation hypervectors targeted at that word $w$. One pass through all training documents, $L^{(T)}_w$ reflects the summary of all occurrences where the word $w$ was encountered until time $T$. In particular, all its contexts are compressed (albeit lossy) and recorded, which will be used as the embedding for the word $w$. 

Unlike back propagation, we use each occurrence (of center word $w$) exactly once and there is no need to store the data set for this purpose. The cost-saving for this streaming fashion can be huge. 

Once produced, the context can be recovered. For example, a near-neighbor search, with a probe of $L^{(T)}_w \oslash P_{step}$, will produce the words that most likely appears immediately after the center $w$. From this perspective, prediction for the next word is merely a by-product from the recovery of compressed contexts. 

\subsection{Discussions}

The proposed training also offers greater clarity and diagnostic insights into the underlying model, enabling and empowering trust-worthy AI systems that can function reliably in mission-critical applications. With clear understand of how the model learns and operates, there should not be any hallucination or surprises.

In addition, the updated semantic code for each word $w$ can be immediately used for inference. From architectural point of view, the training and inference is unified and essentially the training burden is amortized into day-to-day inferences. The simplification can obviously bring significant cost reduction and efficiency boost, but most interestingly, this mirrors closely with our own cognitive ability and experience: human infants learn to speak while babbling. 

Unlike the framing of learning as optimization by neural networks, the proposed learning here is better framed as an unsupervised clustering, where similar observations forms its own clusters. However, supervised learning is also possible under this model, and it's being actively investigated by the author.

Heralding a dramatic departure from the traditional wisdom, this example offers a glimpse into what \textit{damage} this novel cognitive model (and learner) can bring. We believe the heavy-lifting work still lies on re-thinking and re-architecturing critical pieces in the downstream pipelines, for example, the popular Transform architecture (\cite{Vaswani2017}), which is notorious for its opaque-ness and heavy cost. If we can for example, augment Transform architecture with such sparse binary hypervectors, dramatic change can happen to the whole landscape of NLP overnight.
 
\section{Summary}

In essence, we proposed an embodiment of the VSA model with the use of sparse binary hypervectors, which features:
\begin{itemize}
    \item much-needed transparency, viable for a wider range of trust-worthy and reliable AI applications;
    \item greatly reduced cost and improved flexibility;
    \item high efficiency in terms of storage and computation, suitable for mobile and edge deployments.
\end{itemize}

While exploring cognitive model with sparse binary hypervectors, three general principles are hypothesized for a true intelligent system that can potentially match human cognitive capabilities.

Novel learning algorithms are also developed. In particular, the learning algorithms operate in a streaming fashion: new data, as they becomes available, are used to update the model and in principle never needed at a later time. The algorithms are also online in the sense that the updated model can be immediately available for inference as well.

It's my humble hope that this publication can shed some lights into current AI endeavors with a new perspective on transparency and efficiency, and more compelling business cases can be found for trustworthy AI models.

We've developed a high-performance Python library for the manipulations of these sparse binary hypervectors, which is in the process of packaging and releasing. Please contact the author if interested.

\textbf{The ideas discussed here have been filed as a pending patent. While academic redistribution, explorations and improvements are welcome, please contact the author for commercial use.}

\bibliographystyle{unsrtnat}
\bibliography{references}  %%% Uncomment this line and comment out the ``thebibliography'' section below to use the external .bib file (using bibtex) .

%%% Uncomment this section and comment out the \bibliography{references} line above to use inline references.
% \begin{thebibliography}{1}

% 	\bibitem{kour2014real}
% 	George Kour and Raid Saabne.
% 	\newblock Real-time segmentation of on-line handwritten arabic script.
% 	\newblock In {\em Frontiers in Handwriting Recognition (ICFHR), 2014 14th
% 			International Conference on}, pages 417--422. IEEE, 2014.

% 	\bibitem{kour2014fast}
% 	George Kour and Raid Saabne.
% 	\newblock Fast classification of handwritten on-line arabic characters.
% 	\newblock In {\em Soft Computing and Pattern Recognition (SoCPaR), 2014 6th
% 			International Conference of}, pages 312--318. IEEE, 2014.

% 	\bibitem{hadash2018estimate}
% 	Guy Hadash, Einat Kermany, Boaz Carmeli, Ofer Lavi, George Kour, and Alon
% 	Jacovi.
% 	\newblock Estimate and replace: A novel approach to integrating deep neural
% 	networks with existing applications.
% 	\newblock {\em arXiv preprint arXiv:1804.09028}, 2018.

% \end{thebibliography}

\end{document}